\documentclass[letterpaper,12pt]{article}

\usepackage[utf8]{inputenc}
\usepackage[T1]{fontenc}
\usepackage[english]{babel}
\usepackage{mathtools}
\usepackage{mleftright}
\usepackage[mathscr]{euscript}
\usepackage{dutchcal}
\usepackage{graphicx,url}
\usepackage[left=1in,right=1in,top=1in,bottom=1in,letterpaper]{geometry}

\usepackage{dirtytalk}
\usepackage{dsfont}
\usepackage{setspace}
\usepackage{amssymb}
\usepackage[margin = 1cm]{caption}
\usepackage{sidecap}
\usepackage{float}
\usepackage{amsmath}
\usepackage{amsthm}
\usepackage{color,epstopdf}
\usepackage[noend]{algpseudocode}
\usepackage{hyperref}
\usepackage{algorithm,algpseudocode}
\usepackage{breqn}

\usepackage{epsfig}
\usepackage{tabu}
\usepackage[table,xcdraw]{xcolor}
\usepackage{arydshln}
\usepackage{xcolor}
\usepackage{mathrsfs}
\usepackage{caption}
\usepackage{subcaption}

\DeclarePairedDelimiterX{\expectarg}[1]{[}{]}{%
  \ifnum\currentgrouptype=16 \else\begingroup\fi
  \activatebar#1
  \ifnum\currentgrouptype=16 \else\endgroup\fi
}

\newcommand{\LinesNumbered}{
  \setboolean{algocf@linesnumbered}{true}%
  \renewcommand{\algocf@linesnumbered}{\everypar={\nl}}}%
\let\oldnl\nl
\newcommand{\nonl}{\renewcommand{\nl}{\let\nl\oldnl}}

\newcommand{\innermid}{\nonscript\;\delimsize\vert\nonscript\;}
\newcommand\tab[1][1cm]{\hspace*{#1}}

\newcommand{\activatebar}{%
  \begingroup\lccode`\~=`\|
  \lowercase{\endgroup\let~}\innermid 
  \mathcode`|=\string"8000
}

\algnewcommand{\Inputs}[1]{%
  \State \textbf{Inputs:}
  \Statex \hspace*{\algorithmicindent}\parbox[t]{.8\linewidth}{\raggedright #1}
}
\algnewcommand{\Initialize}[1]{%
  \State \textbf{Initialize:}
  \Statex \hspace*{\algorithmicindent}\parbox[t]{.8\linewidth}{\raggedright #1}
}
\algnewcommand{\Data}[1]{%
  \State \textbf{Data:}
  \Statex \hspace*{\algorithmicindent}\parbox[t]{.8\linewidth}{\raggedright #1}
}

\newtheorem{theorem}{Theorem}

\newtheorem{definition}{Definition}

\graphicspath{{images/}} 

\begin{document}
\title{Topic Analysis for Text with Side Data}

\author{
  Biyi Fang\\
  Department of Engineering Science and Applied Mathematics\\
  Northwestern University\\
  Evanston, IL 60208 \\
  \texttt{biyifang2021@u.northwestern.edu} \\
  \and
  Kripa Rajshekhar\\
  Metonymize\\
  \texttt{kripa@metonymize.com}\\
  \and
  Diego Klabjan \\
  Department of Industrial Engineering and Management Sciences \\
  Northwestern University\\
  Evanston, IL 60208 \\
  \texttt{d-klabjan@northwestern.edu} \\
}

\maketitle



\begin{abstract}
\noindent Although latent factor models (e.g., matrix factorization) obtain good performance in predictions, they suffer from several problems including cold-start, non-transparency, and suboptimal recommendations. In this paper, we employ text with side data to tackle these limitations. We introduce a hybrid generative probabilistic model that combines a neural network with a latent topic model, which is a four-level hierarchical Bayesian model. In the model, each document is modeled as a finite mixture over an underlying set of topics and each topic is modeled as an infinite mixture over an underlying set of topic probabilities. Furthermore, each topic probability is modeled as a finite mixture over side data. In the context of text, the neural network provides an overview distribution about side data for the corresponding text, which is the prior distribution in LDA to help perform topic grouping. The approach is evaluated on several different datasets, where the model is shown to outperform standard LDA and Dirichlet-multinomial regression (DMR) in terms of topic grouping, model perplexity, classification and comment generation.\\
\noindent \textbf{Keywords.} LDA, NN
\end{abstract}

\section{Introduction}
\tab As the conjoint and massive knowledge in the forms of news, blogs, web pages, etc., continues to be digitized and stored, discovery becomes more and more challenging and together with it the main underlying topics. A Bayesian multinomial mixture model, latent Dirichlet allocation (LDA) \cite{blei2003latent}, has recently gained much popularity due to its simplicity, and usefulness in stratifying a large collection of documents by projecting every document to a low dimensional space which is spanned by a set of bases capturing the semantic aspects of the collection. However, as the acquisition of information becomes more convenient, text data can be accompanied by extra side data. \textcolor{black}{For example, when customers post their comments for products or restaurants, they usually associate a comment with a rating score or a thumb-up or thumb-down opinion, and retailers usually provide categorical labels for the products in question. In addition, customer loyalty data can be pulled in.} Taking such side data into account improves the ability of LDA to discover patterns and topics among the documents. Currently, there are two types of existing models that combine side data: (1) downstream topic models and (2) upstream topic models. The downstream models assume that the text content and side data are generated simultaneously given latent topics, while upstream models assume that the text content is generated conditioned on the side data as well. Our model belongs to the family of upstream topic models where we accommodate much more complex interactions between side data and text by means of deep neural networks, when compared to other upstream topic models,  i.e. DMR \cite{Mimno2008TopicMC}.

In this paper, we propose a new LDA-style topic model, namely hybrid neural network LDA (nnLDA), based on LDA and a neural network. Our model captures not only text content of the dataset, but also useful \textcolor{black}{topic-level} content and secondary, non-dominant, and more salient statistical patterns from side data. Formally, the model represents the document-topic distribution as mixtures of feature-specific distributions. The prior distribution over topics is the output of a neural network whose input is side data, therefore, it is specific to each distinct combination of side data. Moreover, the neural network is optimized together with the rest of the model in a stochastic EM sampling scheme to better interpret the collection of the documents. The expectation step corresponds to finding the optimal word group and topic group while the maximization step aims to find the optimal neural network parameters and the topic-word distribution. In standard LDA, the prior is fixed while in nnLDA, it depends on a sample.

In this paper, we not only propose a more general model, nnLDA, but also present a complete technical proof confirming that nnLDA performs at least as well as plain LDA in terms of log likelihood. Furthermore, we provide an efficient variational EM algorithm for nnLDA. Lastly, we demonstrate our approach on a few real-world datasets. In summary, we make the following contributions.
\begin{itemize}
    \item We provide a new topic model for text datasets with side data.
    \item We prove that the lower bound of log likelihood of nnLDA is greater than or equal to the lower bound of log likelihood of LDA for any dataset.
    \item We provide an efficient variational EM algorithm for nnLDA.
    \item We present numerical results showing that nnLDA outperforms LDA and DMR in terms of topic grouping, model perplexity, classification and text generation.
\end{itemize}
The paper is organized as follow. In the next section, we review several related works about incorporating side data in generative topic models. In Section 3, we present the hybrid neural network model, followed by our analyses of log likelihood of nnLDA and an efficient variational EM algorithm.
In Section 4, we present experimental results comparing nnLDA with plain LDA and DMR. 

\section{Related Work}
There are a large amount of extensions of the plain LDA model, however, a full retrospection of this immense literature exceeds the scope of this work. In this section, we state several kinds of variations of LDA which are most related to our new model and interpret the relationships among them.

\textit{\textbf{LDA:}}
Plain LDA, a probabilistic latent aspect model, has been widely applied on text documents \cite{blei2003latent} and \cite{Wang2020KeywordbasedTM}, images \cite{Li2005ABH}, and network entities \cite{Airoldi2008MixedMS}
on account of its convenience and functionality in reducing the dimensionality of the data and generating interpretable and semantically coherent topics. \textcolor{black}{Besides, to address the problem that any change to the topic model requires mathematically deriving a new inference algorithm, Srivastava and Sutton \cite{srivastava2017autoencoding} use the Logistic-Normal prior to mimic the simplex in latent topic space and proposed the Neural Variational LDA (NVLDA).} LDA is an unsupervised model which is typically constructed on a distinct bag of words from input contents. Nevertheless, in many practical applications, besides the document contents, useful side information can be easily obtained. Furthermore, such side information often provides useful high-level or direct summarization of the content, while it is not directly involved in the plain LDA model to affect topic inference. In contrast, nnLDA incorporates such information into latent aspect modeling by applying a neural network to discover secondary, non-dominant and more salient statistical patterns that may be more interesting and related to the user's goal.

\textit{\textbf{Downstream Topic Models:}} One approach of incorporating side data in generative topic models is to generate both the text content and side data simultaneously given latent topics. More precisely, for this type of models, each hidden topic not only has a distribution over words but also has another distribution over side data. As a consequence, in training, the loss function in optimization-based learning is the joint likelihood of the content and side data. Examples of such \say{downstream} models are correspondence LDA (Corr-LDA) \cite{Blei2003ModelingAD}, mixed-membership model for authorship \cite{Erosheva2004MixedmembershipMO}, Group-Topic model \cite{Wang2005GroupAT}, Topics over Time model (TOT) \cite{Wang2006TopicsOT}, maximum entropy discrimination LDA (MedLDA) \cite{Zhu2012MedLDAMM} \textcolor{black}{and Term-URL model (TUM) \cite{jiang2013beyond}}. \textcolor{black}{TUM is an application of LDA model onto search engine query log, which captures the characteristics of query terms and URLs separately based on the topic. Consequently, TUM model does not interchange information between the query and the URL. Furthermore, due to the separate generative processes for both the query and the URL, TUM model is computationally demanding.} Another one of the most flexible downstream models is the supervised LDA (sLDA) model \cite{Blei2007SupervisedTM}, which has variants including multi-class sLDA \cite{Wang2009SimultaneousIC} and TOT \cite{Wang2006TopicsOT}. sLDA generates side data such as customers' ratings by maximizing the joint likelihood of the content data and the responses, where the likelihood-based objective is a generalized linear model (GLM) incorporating a proper link function with an exponential family dispersion function specified by the modeler for different types of side data. Compared to our model, in order to explicitly estimate probability distributions over all different side data, sLDA has to fully specify the link function and dispersion functions for the GLM, which increases the modeling complexity significantly. In essence, only a relatively small number of distinct side data vectors are allowed. Our model has no such restriction by applying a completely different approach.

\textit{\textbf{Upstream Topic Model:}}
In a \say{downstream} model, the side data is predicted based on the latent topics of the dataset, whereas in an \say{upstream} topic model, the side data is being conditioned on to generate the latent topics of the dataset. Another distinct difference from \say{downstream} topic models is the choice of the likelihood-based loss in optimization-based learning. More precisely, instead of maximizing the joint likelihood of the content and side data, an \say{upstream} topic model maximizes the conditional likelihood. Examples of such \say{upstream} topic models are Discriminative LDA (DiscLDA) \cite{LacosteJulien2008DiscLDADL}, the scene understanding models \cite{Sudderth2005LearningHM} and the author-topic model \cite{RosenZvi2004TheAM}. In the author-topic model, words are generated by first selecting one author uniformly from an observed author list and then selecting a topic from the topic distribution with respect to that specific author. Then, given a topic, words are selected from the topic-word distribution of that topic. This model assumes that each word is generated only by one author. There are a few extensions of the author-topic model which allow a mixture of latent topics for one document and one author, i.e. \cite{RosenZvi2004TheAM}, \cite{McCallum2007TopicAR}, \cite{Dietz2007UnsupervisedPO}. However, these aforementioned models cannot accommodate combinations of modalities of side data, for example, the aforementioned models cannot handle categorical data and continuous data at the same time. In addition, the side data used in these models are either ratings or labels which essentially is the final intention of learning. Different from the previous models, DMR can handle combinations of modalities of side data, which uses the dot product to project the impact of side data onto the prior. An advanced version introduced in \cite{benton2016collective} is collective supervision of topic models, where aggregate-level labels are provided for groups of documents instead of individual documents, followed by a deterministic relationship between the labels and the priors. Similar to DMR, although the advanced version in \cite{benton2016collective} uses group-level labels and can handle missing side data, it also employs dot product to directly project the impact of the side data onto the prior. Compared to DMR and collective supervision of topic models, nnLDA provides a more comprehensive and flexible learning of side data by applying a neural network when compared to the dot product employed in DMR and the normal distribution assumption in collective supervision of topic models.

\section{Model and Algorithm}
We first present notation and the setting. We use the language of text collections throughout the paper, referring to terminologies such as \say{words,} \say{documents} and \say{corpus} since it makes the concepts more intuitive to understand. 
In general, similar to plain LDA, nnLDA is not restricted to text datasets, and can also be applied on other kinds of datasets, i.e. image datasets. 
\begin{itemize}
\item A $\textit{word}$, defined as an item from a vocabulary indexed by $\left\{1,\cdots,V \right\}$, is applied one-hot encoding. 
More precisely, using superscripts to denote components, the $v$'th word in the vocabulary is represented by a $V$-vector $w$ such that $w^v=1$ and $w^u=0$ for all $u\neq v$.
\item A $\textit{document}$ is a set of $N$ words denoted by $d=\textbf{w}=\left\{w_1,w_2,\cdots,w_N\right\}$ if it only contains textual data
. Similarly, if a document contains $q$ different kinds of side 
data together with the aforementioned textual data, we denote it by $d=(\textbf{w},\textbf{s})=\left(\left\{w_1,w_2,\cdots,w_N\right\}, \left(s_1,s_2,\cdots,s_q\right)\right)$ where $\textbf{s}\in\mathrm{R}^q$.
\item A $\textit{corpus}$ is a collection of $M$ documents denoted by $D=\left\{\textbf{w}_1, \textbf{w}_2,\cdots, \textbf{w}_M\right\}$ for textual only documents and $(D,S)=\left\{(\textbf{w}_1,\textbf{s}_1), (\textbf{w}_2,\textbf{s}_2),\cdots, (\textbf{w}_M,\textbf{s}_M)\right\}$ for documents containing both 
side and textual data.
\end{itemize}

The main goal of nnLDA is to find a probabilistic model of a corpus that, by involving high-level summarization from side data, not only assigns high probability to documents in this corpus but also assigns high probability to other similar documents based on side data.
\subsection{Generative Model}
We propose the nnLDA model to explain the generative process of a document $d$ with textual data $\textbf{w}$ (containing $N$ words) and side data (structural data) $\textbf{s}$, the steps of which can be summarized as follows.
\begin{enumerate}
\item Choose $N\sim\textsf{Poisson}(\xi)$\label{S1}.
\item Choose $\textbf{s}\sim \mathscr{N} (\mu,\sigma^2I)$\label{S2}.
\item Choose $\alpha_d =g(\gamma;\textbf{s})$\label{generateAlpha}
\item Choose $\theta\sim\textsf{Dir}(\alpha_d)$.\label{S4}
\item For each of the $N$ words $w_n$:
\begin{enumerate}
\item Choose a topic $z_n\sim\textsf{Multinomial}(\theta)$.\label{S5a}
\item Choose a word $w_n$ from $p(w_n\mid z_n,\beta)$, a multinomial probability conditioned on the topic $z_n$.\label{S5b}
\end{enumerate}
\end{enumerate}
Notation \say{Poisson,} \say{Dir} represents the Poisson and Dirichlet distribution, respectively. In step \hyperref[generateAlpha]{3}, $g$ refers to a parametric model to generate $\alpha$. In summary, the model has two trainable parameters: $\gamma$, the parameters of $g$ for side info $\textbf{s}$; $\beta$, the topic-word distribution. In the meanwhile, there are three hyper parameters: $\mu$ and $\sigma^2$, the mean and the variance of the probability distribution for side data $\textbf{s}$; and $K$, which does not explicitly appear in the generative process, the number of topics. 

Step \hyperref[S1]{1} is independent of the remaining steps, which determines the number of words in the document. Then, for each document, step \hyperref[S2]{2} provides a representation of side data $\textbf{s}$ by using a normal distribution with mean $\mu$ and variance $\sigma^2$. Then, applying a model with input $\textbf{s}$ in step \hyperref[generateAlpha]{3} provides the prior $\alpha_d$ for the Dirichlet distribution. \textcolor{black}{In step \hyperref[generateAlpha]{3}, the model $g(\gamma;\cdot)$ employed is a deep neural network, and we do not specify the architecture of the deep neural network in this study since different kinds of side data may inquire different deep neural networks. We leave the freedom of selecting the architecture of the deep neural network to the user.} Next, the random parameter of a multinomial distribution over topics, $\theta$, is generated by the Dirichlet distribution. Finally, for the $n$'th word in the document, step \hyperref[S5a]{5(a)} first selects a topic $z_n$ among the $K$ different topics by the multinomial distribution with parameter $\theta$, and then step \hyperref[S5b]{5(b)} generates a word $w_n$ based on the topic-word distribution $\beta$ specific to topic $z_n$. Step \hyperref[S5a]{5} follows standard LDA.
\subsection{Analysis}
Note that a Dirichlet random vector $\theta=\left(\theta_1,\theta_2,\cdots,\theta_K\right)$ has the following probability density:
\begin{align*}
p(\theta\mid\alpha)=\frac{\Gamma\left(\sum_{i=1}^K\alpha_i \right)}{\prod_{i=1}^K\Gamma(\alpha_i)}\theta_1^{\alpha_1-1}\cdots\theta_K^{\alpha_K-1},
\end{align*}
where $K$ is the number of topic groups, $\alpha$ is the prior of the Dirichlet distribution and $\theta$ takes values in the $(K-1)$-simplex. Then, the generative process implies that the conditional distribution of the nnLDA model of a document $d=(\textbf{w},\textbf{s})$ is
\begin{align*}
P_1(\textbf{w}\mid \mu,\sigma,\gamma,\beta)
&=\tilde{\tilde{P}}_1(\textbf{w}\mid \textbf{s},\gamma,\beta)\\
&=\int \tilde{\tilde{p}}(\theta\mid\textbf{s},\gamma)\left(\prod_{n=1}^{N}\sum_{z_{k}}\tilde{p}(z_{k}\mid \theta)\tilde{p}(w_{n}\mid z_{k},\beta)\right)\mathrm{d}\theta\\
&=\int \tilde{p}(\theta\mid\mu,\sigma,\gamma)\left(\prod_{n=1}^{N}\sum_{z_{k}}\tilde{p}(z_{k}\mid \theta)\tilde{p}(w_{n}\mid z_{k},\beta)\right)\mathrm{d}\theta\\
&=\int \tilde{p}(\theta\mid\mu,\sigma,\gamma)\left(\prod_{n=1}^{N}\sum_{i=1}^{K}\prod_{j=1}^V (\theta_i\beta_{ij})^{w_n^j} \right)\mathrm{d}\theta,
\end{align*}
which in turn yields
\begin{align*}
P_1(D\mid\mu,\sigma,\gamma,\beta)
&=\mathbb{E}\left[\int \tilde{p}(\theta_d\mid \mu,\sigma, \gamma)\left(\prod_{n=1}^{N}\sum_{z_{d_k}}\tilde{p}(z_{d_k}\mid \theta_d)\tilde{p}(w_{d_n}\mid z_{d_k},\beta)\right)\mathrm{d}\theta_d\right]\\
&=\mathbb{E}\left[\int \tilde{p}(\theta_d\mid \mu,\sigma, \gamma)\left(\prod_{n=1}^{N}\sum_{i=1}^{K}\prod_{j=1}^V (\theta_i\beta_{ij})^{w_n^j} \right)\mathrm{d}\theta_d\right],
\end{align*}
where $\tilde{p}(\theta_d\mid\mu,\sigma,\gamma)=\tilde{\tilde{p}}(\theta_d\mid\textbf{s},\gamma)=p(\theta_d\mid g(\gamma;\textbf{s}))=p(\theta_d\mid\alpha_d)$ for a corpus $D$. 

The nnLDA model represented above is a probabilistic graphical model with three levels. Parameters $\mu$, $\sigma$, $\gamma$ and $\beta$ are corpus-level parameters, which are assumed to be sampled once in the generative process of a corpus. Variables $\alpha_d$ and $\theta_d$ are document-level variables, which are sampled once per document. Finally, $w_{d_n}$ and $z_{d_k}$ are word-level variables, sampled once for each word in each document.

In the rest of this section, we provide an analytical comparison of standard LDA and nnLDA.

Compared to standard LDA, nnLDA employs an extra neural network $g$ to generate document-level variable $\alpha_d$. Since nnLDA is \say{richer} than LDA, we expect that it should produce a higher likelihood. Without assumptions on $g(\gamma;\cdot)$ this does not hold since, for example, $g(\gamma;\cdot)$ can map everything to a constant vector different from the prior used by LDA. As a result, in order for the statement to hold the network must be expressive. The question to consider is whether a neural network is capable of memorizing arbitrary side data of a given size. We tackle this question by introducing the concept of finite sample expressivity which is an extension of a similar definition in \cite{Yun2019SmallRN}. Given the definition, if $g(\gamma;\cdot)$ has finite sample expressivity, nnLDA at least can find the optimal $\alpha^*$ used in standard LDA. 

\begin{definition}
\label{def:1}
Function $g(\gamma;\cdot)$ has finite sample expressivity if for all inputs $x_i\in\mathrm{R}^{d_x}, 1\leq i\leq N$ and for all $y_i\in[-M,+M]^{d_y}, 1\leq i\leq N$ for some constant $M > 0$, there exists a parameter $\gamma$ such that $g(\gamma;x_i)=y_i$ for every $1\leq i\leq N$.
\end{definition}\label{def:1}
Based on Definition \hyperref[def:1]{1}, Theorem 3.1 shown in \cite{Yun2019SmallRN} provides a specific set of constraints, i.e. any 3-layer (i.e., 2-hidden-layer) ReLU FCNN with hidden layer widths $d_1$ and $d_2$ can fit any arbitrary dataset if $d_1d_2 \geq 4Nd_y$, where $d_y$ and $N$ are the dimension of the label and the number of samples, respectively. By extending the aforementioned theorem, Proposition 3.4 and Theorem 4.1 in \cite{Yun2019SmallRN} argue that any FCNN given constraints on the number of neurons in each layer is able to have finite sample expressivity. 

In the following, we assume that $g(\gamma;\cdot)$ has finite sample expressivity. Therefore, given $K$ and any $\alpha^*$ representing the number of topic groups and optimal parameters in LDA, since $\alpha^*\in [-M, +M]^K$ for some constant $M$, there exists a $\gamma_1$ such that, for all inputs $\textbf{s}_i$ and $\alpha^*$, $g(\gamma_1;\textbf{s}_i)=\alpha^*$ for all $1\leq i \leq N$. 

We next prove that the optimized probability of nnLDA is at least as good as that of plain LDA. Let $\alpha^*$ and $\beta^*$ be optimal solutions to $P_2 = \max_{\alpha,\beta} P(\left.D\right|\alpha, \beta)$  of LDA, meanwhile, let $\mu^*, \sigma^*$ and $\gamma ^*$ be optimal solutions to $P_1=\max_{\mu,\sigma,\gamma}P_1(\left.D\right|\mu,\sigma,\gamma,\beta^*)$ of nnLDA (see Appendix \hyperref[eq:lda]{A} for formal definitions).

\begin{theorem}\label{thm:1}
If $\alpha^*$, $\beta^*$ are optimal solutions to LDA, then there exists optimal solutions $\mu^*,\sigma^*$ and $\gamma^*$ to nnLDA such that
\begin{align*}
P_1(D\mid \mu^*,\sigma^*,\gamma^*,\beta^*)\geq P_2(D\mid\alpha^{*},\beta^{*}).
\end{align*}
\end{theorem}
\begin{proof}
See Appendix \hyperref[pf:thm1]{B}.
\end{proof}

While Theorem \ref{thm:1} asserts that when it comes to model fit nnLDA fits the data better than LDA, it does not provide a gap statement. If the side data provides positive influence during the learning process by a constant $C$, then, due to the independence of words, topics and documents, we are able to argue that the optimized probability is at least improved by $C-1$.
\begin{theorem}\label{thm:2}
For any document $(\textbf{w},\textbf{s})\in (D,S)$, if $\hat{p}(w_i\mid\alpha^*, \beta^*)\neq 0$ for all $i$, and there exists a positive constant $C>1$ such that $\prod_{i=1}^N\tilde{p}(w_i\mid \gamma^*, \beta^*, \mu^*,\sigma^*)\geq C\prod_{i=1}^N\hat{p}(w_i\mid\alpha^*, \beta^*)$ for every $w_i\in \textbf{w}$, and if $D$ in $P_1$ and $D$ in $P_2$ follow the same distribution, then
\begin{align*}
\frac{P_1(D\mid\mu^*,\sigma^*,\gamma^*,\beta^* )-P_2(D\mid\alpha^*,\beta^*)}{P_2(D\mid\alpha^*,\beta^*)}\geq C-1.
\end{align*}
\end{theorem}
\begin{proof}
See Appendix \hyperref[pf:thm2]{C} for a formal proof.
\end{proof}
The assumption on $\hat{p}(w_i\mid\alpha^*, \beta^*)$ in Theorem \ref{thm:2} is reasonable since it indicates that all documents are not randomly generated. 
The positive constant $C$ in the assumption captures the improvement given by the side data. In other words, as long as the side data has positive impact on the text data, this assumption holds. Next, we link the existence of $C$ to lift from data mining.
Let us define lift as
\begin{align*}
l(d) = \frac{P(\textbf{w})P(\textbf{s})}{P(\textbf{w}, \textbf{s})}
\end{align*}
with $d=(\textbf{w},\textbf{s})$. Lift measures the dependency level of words $\textbf{w}$ and side data $\textbf{s}$. If $l(d) < 1$ for $d$ with $N$ words and $P(\textbf{s})>0$, we have 
\begin{align*}
P(\textbf{s})\prod_{n=1}^NP(w_{n}) = P(\textbf{w})P(\textbf{s}) &< P(\textbf{w}, \textbf{s}) =P(\textbf{s})\prod_{n=1}^NP(w_{n}|\textbf{s}),
\end{align*}
and in turn
\begin{align*}
\prod_{n=1}^NP(w_{n}) & < \prod_{n=1}^NP(w_{n}|\textbf{s}),
\end{align*}
and
\begin{align*}
\prod_{n=1}^N\hat{p}(w_{n}\mid\alpha^*, \beta^*)&< \prod_{n=1}^N\tilde{p}(w_{n}\mid \gamma^*, \beta^*, \mu^*,\sigma^*).
\end{align*}
This implies that there exists $C>1$. In summary, when $l(d) < 1$ and $P(\textbf{s}) > 0$ for each $d$ in the corpus, Theorem \ref{thm:2} holds. Lift essentially measures the dependency of $\textbf{w}$ and $\textbf{s}$, which is widely used in data mining. The condition indicates that the side data helps to link the words to the documents they are more likely to be in.

Informally, in the proof, due to the independence assumption of words, topics and documents in nnLDA, the generative probability of nnLDA for a corpus can be reformulated as a product of $\tilde{p}(\theta_d\mid\mu^*,\sigma^*,\gamma^*)$ and conditional probability of words $\tilde{p}(w_{n}\mid\theta_d,\beta^*)$. Likewise, the same property holds for plain LDA. Lastly, given a relationship between documents $d=\textbf{w}$ and $d=(\textbf{w},\textbf{s})$ as an expression of the conditional probability of words, we are able to build a connection of the optimized probabilities between nnLDA and LDA.
\subsection{Variational Inference with EM Algorithm}
We train the nnLDA model using a stochastic EM sampling scheme, in which we alternate between sampling topic assignments from the current prior distribution conditioned on the observed words and side data, and optimizing the parameters given the topic assignments.

Details are similar to those in \cite{blei2003latent}. In this section, instead of showing all the details, we only point out the differences from the derivation of plain LDA. By applying the Jensen's inequality and KL divergence between the variational posterior probability and the true posterior probability, which is a formally stated technique in \cite{blei2003latent}, a lower bound of log likelihood reads
\begin{align}
    L(\xi,\phi;\gamma,\beta)&=\mathbb{E}_q\left[ \log p(\theta\mid g(\gamma;\textbf{s}))\right]+\mathbb{E}_q\left[ \log p(z\mid \theta)\right]+\mathbb{E}_q\left[ \log p(\textbf{w}\mid z,\beta)\right]\nonumber\\
    &-\mathbb{E}_q\left[ \log q(\theta)\right]-\mathbb{E}_q\left[ \log q(z)\right],
    \label{loss}
\end{align}
where $\xi, \phi$ are variational parameters of $\theta$ and $z$, respectively, and $q(\cdot)$ represents the variational distribution. Then, the iterative algorithm is
\begin{enumerate}
    \item (E-step) For each document, find the optimizing values of the variational parameters $\xi$ and $\phi$ of $z$ and $\theta$, respectively.
    \item (M-step) Maximize the resulting lower bound of log likelihood with respect to the model parameters $\gamma$ and $\beta$.  
\end{enumerate}
The E-step 
is similar to the E-step in \cite{blei2003latent} except replacing prior $\alpha$ by $g(\gamma;\textbf{s})$. We run the E-step until it converges for each document. The M-step is finding a maximum likelihood estimation with expected sufficient statistics for each document under the approximate posterior parameters $\xi$ and $\phi$, which are computed in the E-step. Likewise, since the log likelihood objective related to $\beta$ does not involve $g(\gamma;\textbf{s})$, we are allowed to directly borrow the update rule of $\beta$ from \cite{blei2003latent}, which is
\begin{align*}
    \beta_{ij}\propto \sum_{d=1}^M\sum_{n=1}^{N}\phi^*_{dni}w_{dn}^j.
\end{align*}
 In contrast, for the neural network parameter $\gamma$, we resort to log likelihood objective related to $\gamma$ as follows,
 \begin{align*}
     L_{[\gamma]}=&\sum_{d=1}^M\left(\log \Gamma(\sum_{j=1}^K [g(\gamma;\textbf{s}_d)]_j)-\sum_{i=1}^K\log \Gamma([g(\gamma;\textbf{s}_d)]_i)\right.\\
     &\left.+\sum_{i=1}^K\left(\left([g(\gamma;\textbf{s}_d)]_i-1\right)\left(\Psi(\xi_{di})-\Psi(\sum_{j=1}^K\xi_{dj})\right) \right)
     \right),
 \end{align*}
 where $M$ is the number of documents in the corpus, and $\Psi$ is the digamma function, the first derivative of the log Gamma function. Then, applying the backpropagation approach provides the derivative and the update rule for parameter $\gamma$.
\section{Experimental Study}
In this section, we compare the nnLDA model with standard LDA and the DMR model introduced in \cite{blei2003latent} and \cite{Mimno2008TopicMC}, respectively. We conduct experiments on five different-size datasets among which one is a synthetic dataset and the remaining four are real-world datasets. For these datasets, we study the performance of topic grouping, perplexity, classification and comment generation for nnLDA, plain LDA and DMR models. For each of the tasks, some datasets are not eligible to be examined due to lack of information. The synthetic dataset is publically available at \url{https://github.com/biyifang/nnLDA/blob/main/syn_file.csv} while the real-world datasets are proprietary.
\subsection{Datasets and Training Details}
The first dataset we use is a synthetic dataset of 2,000 samples. Each sample contains a customer’s feedback with respect to his or her purchase along with the characteristics of the product. More precisely, there are two different categories, which are product and description. In the product category, it can either be TV or burger; similarly, in the description category, the word can either be price or quality. In order to generate comments, we assign a bag of words to each combination of product and description as shown in Table \ref{tb1:syn}. 
\begin{table}[H]
\begin{tabular}{|l|l|}
\hline
Category combination & \multicolumn{1}{c|}{Bag of words}                                                                                                   \\ \hline
(burger, price)      & \begin{tabular}[c]{@{}l@{}}value, pricey, ouch, steep, cheap, value, reason, accept, \\unreason, unacceptable\end{tabular}         \\ \hline
(burger, quality)    & \begin{tabular}[c]{@{}l@{}}nasty, fantastic, delicious, tasty, juicy, unreason, unacceptable, \\reason, accept, fresh\end{tabular} \\ \hline
(TV, price)          & \begin{tabular}[c]{@{}l@{}}promotion, affordable, value, increase, expensive, tasty, \\economical, fancy, okay\end{tabular}         \\ \hline
(TV, quality)        & \begin{tabular}[c]{@{}l@{}}fabulous, fantastic, promising, sharp, large, clear, eco friendly, \\fresh, pixilated\end{tabular}      \\ \hline
\end{tabular}
\caption{Synthetic Dataset}\label{tb1:syn}
\end{table}
After randomly selecting one category combination from the four combinations, a comment is generated containing at least one word and at most five words with an average 2.97 words, by selecting a certain number of words at random from the corresponding bag. 

The second dataset is a real-world dataset, PTS for short, which has 795 samples. Each sample contains a customer’s short feedback and rating with respect to his or her purchase along with the characteristics of the product. Additionally, the category (side data) selected for nnLDA corresponds to sectors, which are generalizations of products. In this dataset, there is only only 1 word in the shortest comment, while the longest comment in the dataset contains 49 words. Overall, the average length of the comments is 10.6 words. For example, a customer, who bought a product belonging to sector Baby, leaves a comment \say{Cheap\& Soft} with a rating of 3.

The third dataset WIP is a medium-size dataset with 3,451 samples. Each sample contains a customer’s short feedback and rating with respect to his or her purchase along with the characteristics of the product. The sector attribution is again side data when training models with one feature. The other attribution counted for models with two features is channel. The most concrete comment in the dataset has 138 words, while the briefest comment has only 1 word. In the meanwhile, the average length of the comments in the dataset is 8.9 words.

DCL is another medium-size dataset of 5,427 samples. Different from the PTS and WIP datasets, each sample in DCL contains a customer’s long feedback and rating with respect to his or her purchase along with the characteristics of the product. Additionally, the side data selected for nnLDA corresponds to groups of products. The smallest number of words for a comment in this dataset is 1, while the largest is 988. Overall, the average length of the comments is 61.7 words. A short sample comment is \say{quick points that will be all that matters to a buyer wanting accurate metrics to buy by tinny sound but plenty of audio hookups.}

The last dataset is RR, which has 100,000 samples, from which we randomly select 10,000 samples. Each sample contains a customer’s feedback with respect to his or her purchase along with the characteristics of the product. Additionally, the side data for nnLDA corresponds to the category, which can be grocery, health and personal care, furniture, kitchen, etc. The longest comment has 418 words while the shortest comment has only 1 word as the previous datasets, and the average length of the comments is 69.5 words. A short sample comment reads \say{great tasting oil and made the most excellent gluten free chocolate cake.}

Due to the lack of some information from the certain datasets, we are unable to study all tasks of interest for all of these datasets. For the topic grouping task, we examine the ability of nnLDA, plain LDA and DMR to assign the comments from the same topic group into the same correct topic group. For this task, we only conduct experiments on the synthetic dataset since only the topic groups of the synthetic dataset are clear. For the perplexity task, we compute the logarithm of the perplexity of all the words in the corresponding dataset. We do not study the performance of perplexity for the synthetic dataset since we know the true number of topic groups. For the classification task, we use the probability vector generated by the topic models to predict the rating for that comment. Since the RR dataset does not have ratings, we are unable to examine the classification ability of the topic models on the RR dataset. The last task tests the performance of the topic models on generating new comments. For this task, we only conduct experiments on the two smallest real world datasets since it is of interest how topic models perform given a small number of samples. Table \ref{tb6:tasks} presents the tasks of interest for each dataset.

\begin{table}[H]
\centering
\begin{tabular}{|l|l|l|l|l|}
\hline
Dataset                         & \multicolumn{1}{c|}{Topic grouping} & Perplexity & Classification & \begin{tabular}[c]{@{}l@{}}Comment\\ generation\end{tabular} \\ \hline
Synthetic dataset               & Yes                                 & No         & No           & No  \\ \hline
PTS & No               & Yes        & Yes            & Yes \\ \hline
WIP           & No               & Yes        & Yes            & Yes  \\ \hline
DCL        & No& Yes        & Yes            & No  \\ \hline
RR & No& Yes  & No  & No  \\ \hline
\end{tabular}
\caption{Tasks of Interest}\label{tb6:tasks}
\end{table}
For all of these datasets, we employ a two-layer fully connected neural network as $g(\gamma;\cdot)$ in nnLDA. Furthermore, we set the number of neurons to be 20 in the first layer, the number of neurons of the second layer to be the number of topic groups assigned in the beginning and the batch size to be 64. All features of the side data are categorical and are one-hot encoded. Additionally, all weights in $g(\gamma;\cdot)$ are initialized by Kaiming Initialization \cite{he2015delving}. We apply the ADAM algorithm with the learning rate of $0.001$ and weight decay being $0.1$. Meanwhile, we train all the models using EM with exactly the same stopping criteria of stopping E-step and M-step when the average change over the whole training dataset in the expected log likelihood becomes less than 0.01\%. We vary the number of topic groups from 4 to 30. For DMR, we use the same values for the parameters as those in \cite{Mimno2008TopicMC}. All the algorithms are implemented in Python with Pytorch and trained on a single GPU card. 
\subsection{Experimental Results}
In this section, we present all the results based on the tasks of interest.

Overall, nnLDA outperforms plain LDA and DMR in all datasets in terms of topic grouping, classification, perplexity and comment generation. Meanwhile, based on the fact that the last two datasets have many more words and more intrinsic concepts in their comments when compared to the first three datasets, nnLDA exceeds the performance of plain LDA and DMR dramatically when a document contains several topics or it is more comprehensive.
\subsubsection{Topic Grouping}
Table \ref{tb7:topic} shows the most frequent 5 words in each topic group generated by plain LDA, DMR and nnLDA when setting the number of topic groups to be 4 in the synthetic dataset. The topic groups generated by plain LDA and DMR are very vague and it is very hard to distinguish which topic group is describing what combination of product and description, while the topic groups given by nnLDA are very distinguishable, i.e. topic group 1 is about (burger, quality), topic group 2 is about (TV, price), topic group 3 is about (TV, quality) and topic group 4 is about (burger, price). It identifies correctly the seed topics. Therefore, nnLDA outperforms plain LDA in grouping.
\begin{table}[H]
\begin{tabular}{|l|l|l|l|}
\hline & \multicolumn{1}{c|}{plain LDA}    &DMR        & nnLDA   \\ \hline
Topic group 1 & \begin{tabular}[c]{@{}l@{}}promising, rebate, sharp,\\ increase, outstanding\end{tabular} & \begin{tabular}[c]{@{}l@{}}pricey, unacceptable,\\ juicy, pixilated  \end{tabular} & \begin{tabular}[c]{@{}l@{}}unreason, unacceptable, \\juicy delicious, nasty\end{tabular}   \\ \hline
Topic group 2 & \begin{tabular}[c]{@{}l@{}}unreason, value, okay,\\ steep, ecofriendly\end{tabular}   &\begin{tabular}[c]{@{}l@{}}ouch, steep, tasty,\\ unreason, promotion\end{tabular}       & \begin{tabular}[c]{@{}l@{}}promotion, increase, tasty,\\ economical, okay\end{tabular}     \\ \hline
Topic group 3 & \begin{tabular}[c]{@{}l@{}}reason, accept, promotion,\\ large, unacceptable\end{tabular}  & \begin{tabular}[c]{@{}l@{}}accept, fantastic, value \\reason, affordable\end{tabular}   & \begin{tabular}[c]{@{}l@{}}fresh, promising, fantastic,\\ large, eco friendly\end{tabular} \\ \hline
Topic group 4 & \begin{tabular}[c]{@{}l@{}}fresh, reason, outstanding,\\  ecofriendly, fantastic\end{tabular}&\begin{tabular}[c]{@{}l@{}}sharp, delicious,\\ accept, fresh, clear\end{tabular}  & \begin{tabular}[c]{@{}l@{}}reason, accept, value,\\ steep, cheap \end{tabular} \\\hline
\end{tabular}
\caption{Top words of groups generated by LDA, DMR and nnLDA}\label{tb7:topic}
\end{table}

\begin{table}[H]
\centering
\begin{tabular}{|l|r|r|r|r|}
\hline
& macro-precision & macro-recall & macro-F1&  micro-F1  \\ \hline
LDA        & 0.7238 &0.7272&0.7211& 0.7240\\ \hline
DMR       &0.7238&  0.7460 &     0.7313&  0.7392 \\ \hline
nnLDA     &  0.7401   &  0.7919  &   0.7536 &  0.7905  \\ \hline
relative improvement from LDA &2.25\%&8.90\% & 4.51\%& 9.19\%\\ \hline
relative improvement from DMR &2.25\%& 6.15\%& 3.05\%&6.94\%\\ \hline
\end{tabular}
\caption{Precision, recall and relative improvement of the synthetic dataset generated by LDA, DMR and nnLDA}\label{tb10:micro-macro}
\end{table}

Additionally, based on the top words of topics generated by LDA, DMR and nnLDA, we are able to assign the most related category combination to a comment with respect to a model. Since we have the category combination of each comment, Table \ref{tb10:micro-macro} shows the macro-recall, macro-precision and macro-F1 scores and micro-F1 of LDA, DMR and nnLDA, respectively, when training on the synthetic dataset, and the overall relative improvement of nnLDA. As the table shows, nnLDA outperforms plain LDA and DMR, which implies that nnLDA assigns more samples correctly to the right topic group. Therefore, in general, nnLDA improves the recall, precision and F1 scores. 

In conclusion, nnLDA outperforms standard LDA and DMR in terms of the ability of topic grouping.
\subsubsection{Perplexity}
\begin{figure}[H]
\centering
\begin{minipage}{.5\textwidth}
  \centering
  \includegraphics[width=.95\linewidth]{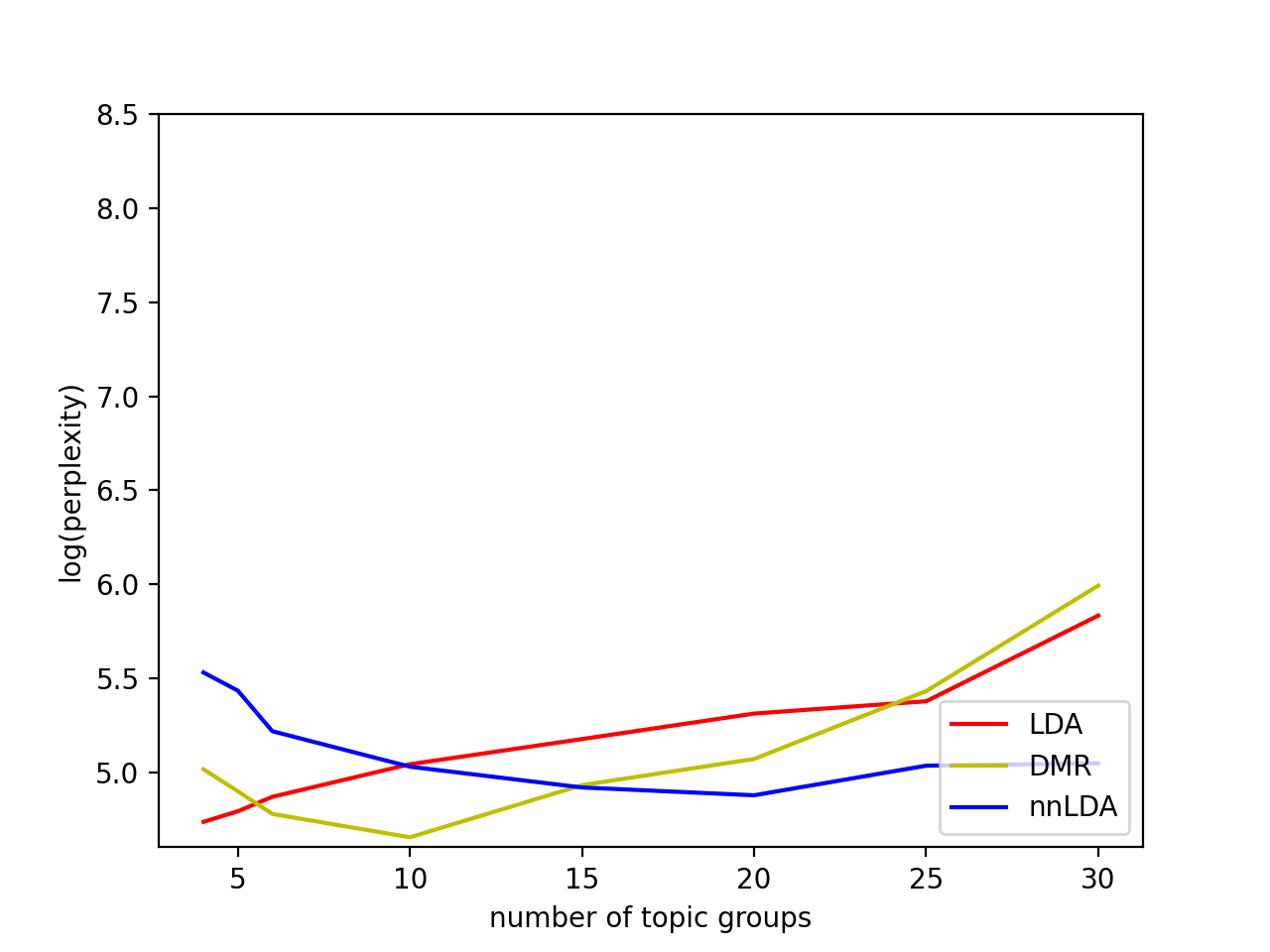}
  \captionof{figure}{PTS dataset}
  \label{fig:proj_perp}
\end{minipage}%
\begin{minipage}{.5\textwidth}
  \centering
  \includegraphics[width=.95\linewidth]{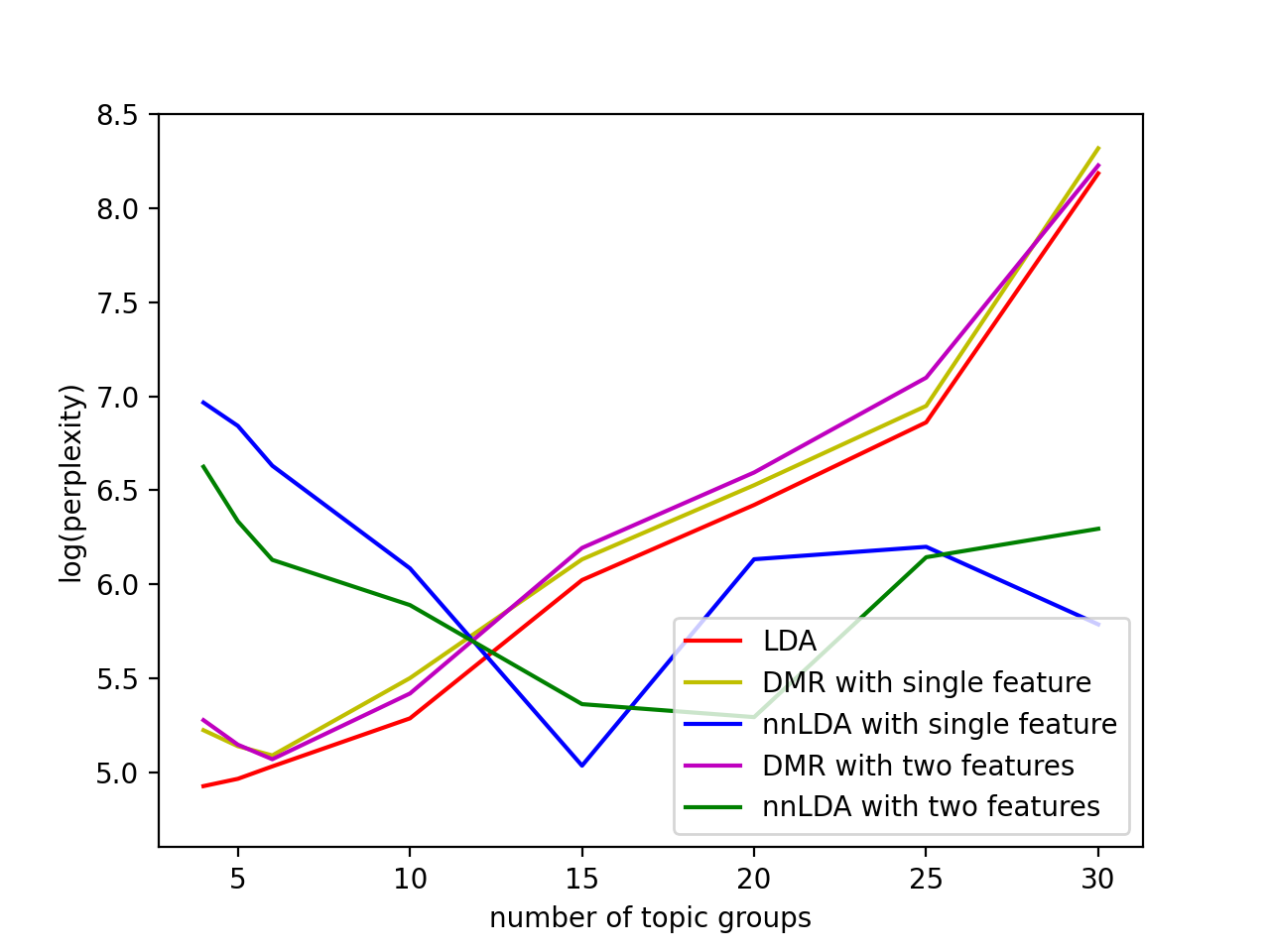}
  \captionof{figure}{WIP dataset}
  \label{fig:whatif_perp}
\end{minipage}
\end{figure}

Figures \ref{fig:proj_perp} and \ref{fig:whatif_perp} represent the log(perplexity) of plain LDA, DMR and nnLDA on the PTS and WIP datasets, respectively. Additionally, in Figure \ref{fig:whatif_perp}, for DMR and nnLDA, we not only conduct experiments on the dataset with the single feature (sector) as the side data, denoted as \say{DMR with single feature} and \say{nnLDA with single feature,} but also on the dataset with two features (sector and channel) as side data, denoted as \say{DMR with two features} and \say{nnLDA with two features,} respectively. The smallest log(perplexity) values generated by plain LDA and DMR are competitive to those of nnLDA for these two datasets. In Figure \ref{fig:proj_perp}, the log(perplexity) value generated by plain LDA increases as the number of topic groups grows, while the log(perplexity) values generated by DMR and nnLDA decrease first and then increase as the number of topic groups increases on the PTS dataset. As it is shown in Figure \ref{fig:whatif_perp}, the log(perplexity) values generated by plain LDA and DMR increase as the number of topic groups grows on the WIP dataset. However, the log(perplexity) values generated by nnLDA decrease first and then increase as the number of topic groups increases on both of the aforementioned datasets. Moreover, we examine DMR and nnLDA models with two features on the WIP dataset, which take both sector and channel attributions as side data into account, in Figure \ref{fig:whatif_perp}. As we can observe, the minimum log(perplexity) generated by nnLDA with two features (sector and channel attributions) is better than that of nnLDA with the single feature (sector attribution), although the optimal number of topic groups occurs at a different point since more side data is provided. Consequently, plain LDA does not learn the datasets, and DMR is able to learn the small datasets. In contrast, nnLDA starts learning the datasets as the log(perplexity) value decreases in the beginning and finds an optimal number of topic groups, then it gets confused since the number of topic groups are more than needed. Furthermore, nnLDA with two features provides better log(perplexity) than nnLDA with the single feature. Therefore, nnLDA is more capable of understanding the datasets; both small and medium-size datasets with short comments. 

\begin{figure}[H]
\centering
\begin{minipage}{.5\textwidth}
  \centering
  \includegraphics[width=.95\linewidth]{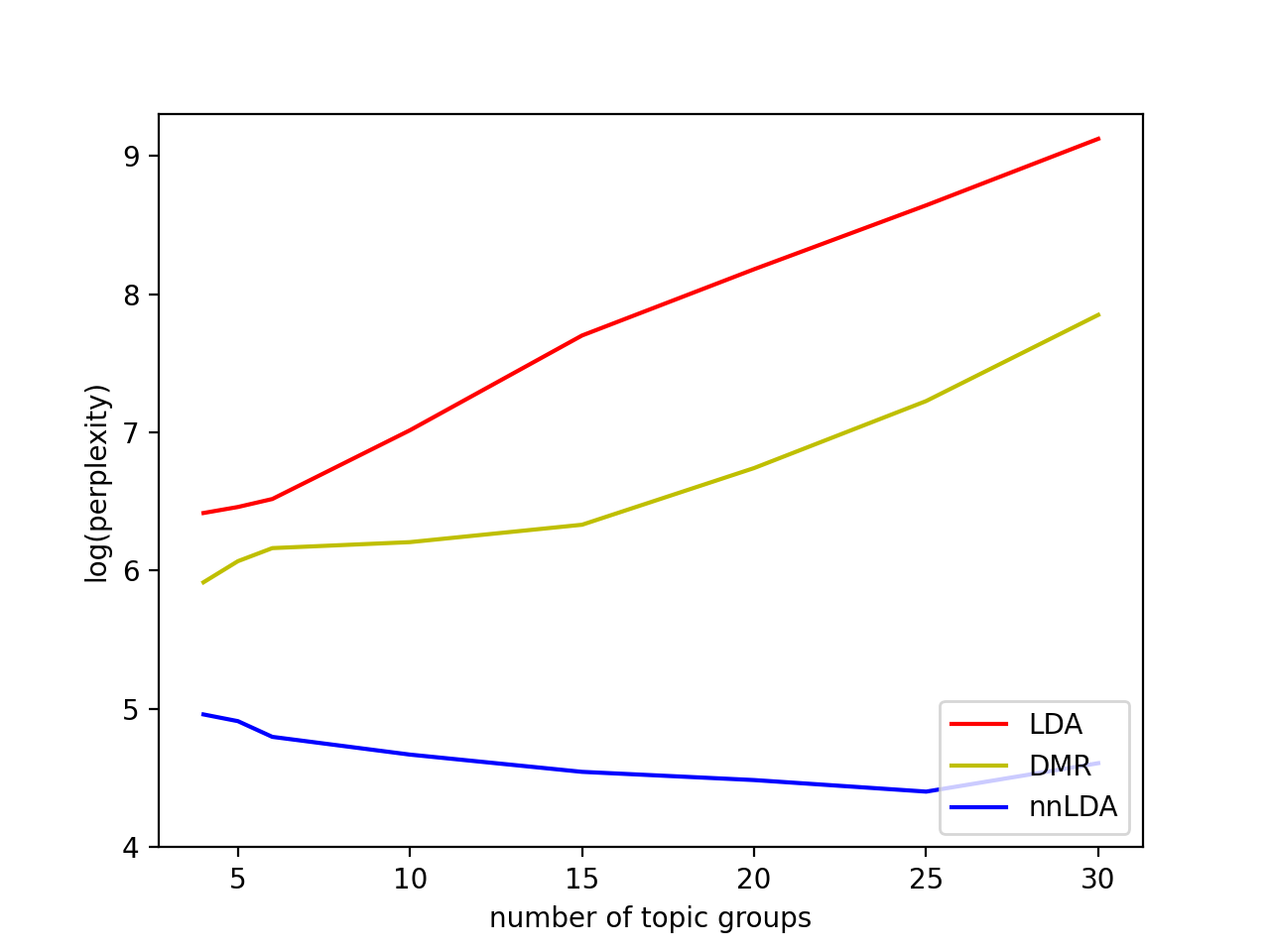}
  \captionof{figure}{DCL dataset}
  \label{fig:TV_2}
\end{minipage}%
\begin{minipage}{.5\textwidth}
  \centering
  \includegraphics[width=.95\linewidth]{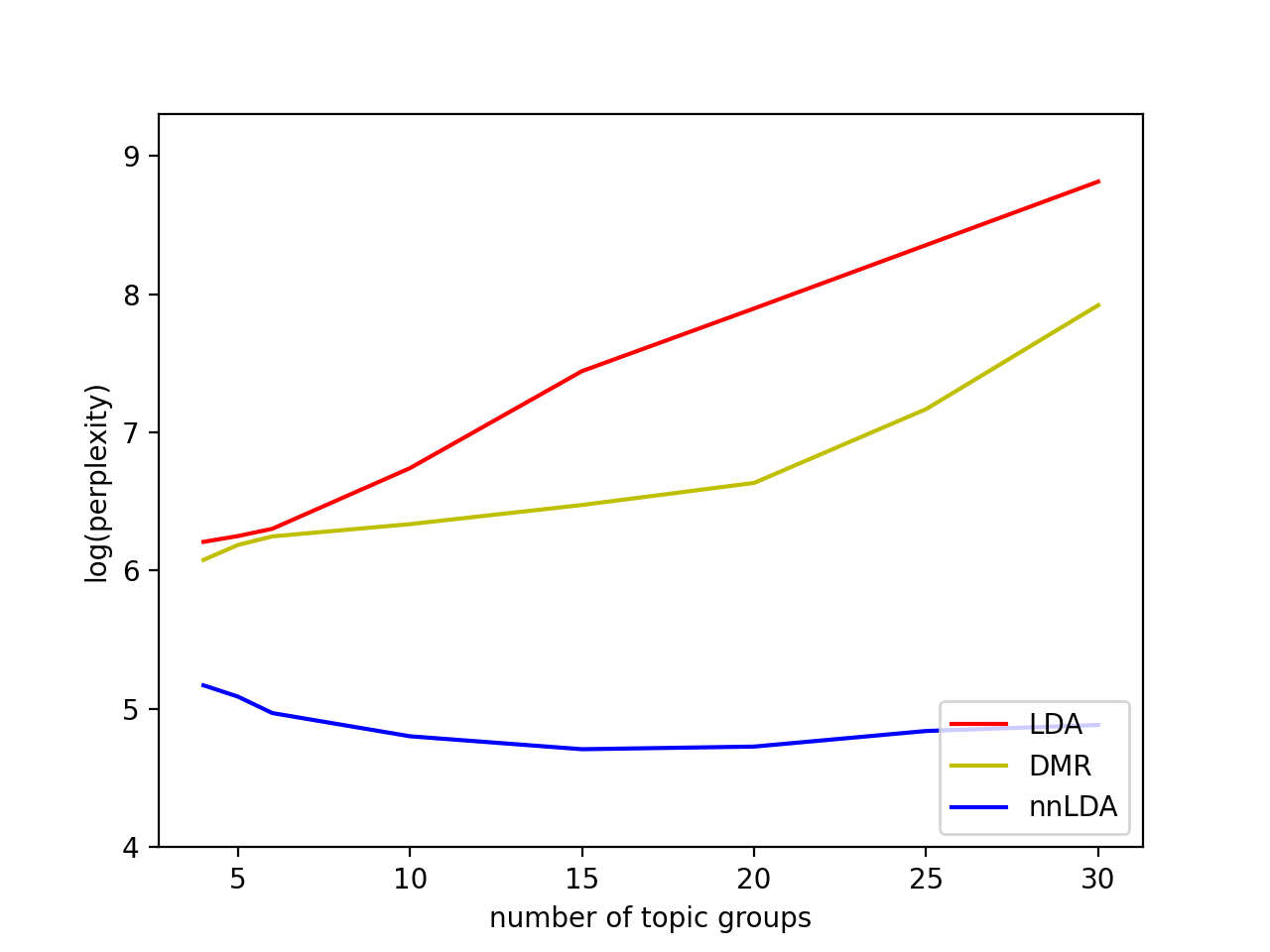}
  \captionof{figure}{RR dataset}
  \label{fig:review_2}
\end{minipage}
\end{figure}

When learning more complex datasets, the advantage of the nnLDA model becomes more pronounced. Figure \ref{fig:TV_2} represents the log(perplexity) of plain LDA, DMR and nnLDA on the DCL dataset, while Figure \ref{fig:review_2} shows the same on the RR dataset. In these figures we observe that the log(perplexity) generated by plain LDA and DMR blows up as the number of topic groups increases, while the log(perplexity) generated by nnLDA decreases first and then increases as the number of topic groups grows. Furthermore, the log(perplexity) values of nnLDA are much smaller than those of plain LDA and DMR. Consequently, nnLDA performs as well as plain LDA and DMR in small and medium-size datasets with short comments, and at the same time, nnLDA explains the datasets better than plain LDA and DMR in medium and large size datasets with long comments. There is also a trade-off between the accuracy and running time as shown in Table \ref{tb:run_time}. In the table, we compare the running time of plain LDA, DMR with one feature and nnLDA with one feature on three different datasets. We observe that nnLDA spends more time than both DMR and plain LDA on training. In conclusion, nnLDA performs better on learning while it requires a slightly longer training time. It is less than $10\%$ slower than DMR.

\begin{table}[H]
\centering
\begin{tabular}{|l|r|r|r|}
\hline
 running time(s) & plain LDA & DMR    & nnLDA  \\ \hline
PTS          & 3      & 4   & 4   \\ \hline
WIF          & 19     & 24  & 26  \\ \hline
DCL          & 138    & 179 & 191 \\ \hline
\end{tabular}
\caption{Running time of different models on different datasets}\label{tb:run_time}
\end{table}

In the following section, we study the classification problem of predicting the rating of each sample. In all the cases, we use 10-fold cross validation, which holds out 10\% of the data for test purposes and trains the models on the remaining 90\%. We apply nnLDA, plain LDA and DMR to find the probability of each sample to be assigned to each topic group and treat it as the feature matrix. Lastly, we train a classification model (xgboost \cite{Chen2016XGBoostAS}) on the feature matrix with the rating labels as the ground truth.
\subsubsection{Classification}
\begin{figure}[H]
\centering
\begin{minipage}{.33\textwidth}
  \centering
  \includegraphics[width=.95\linewidth]{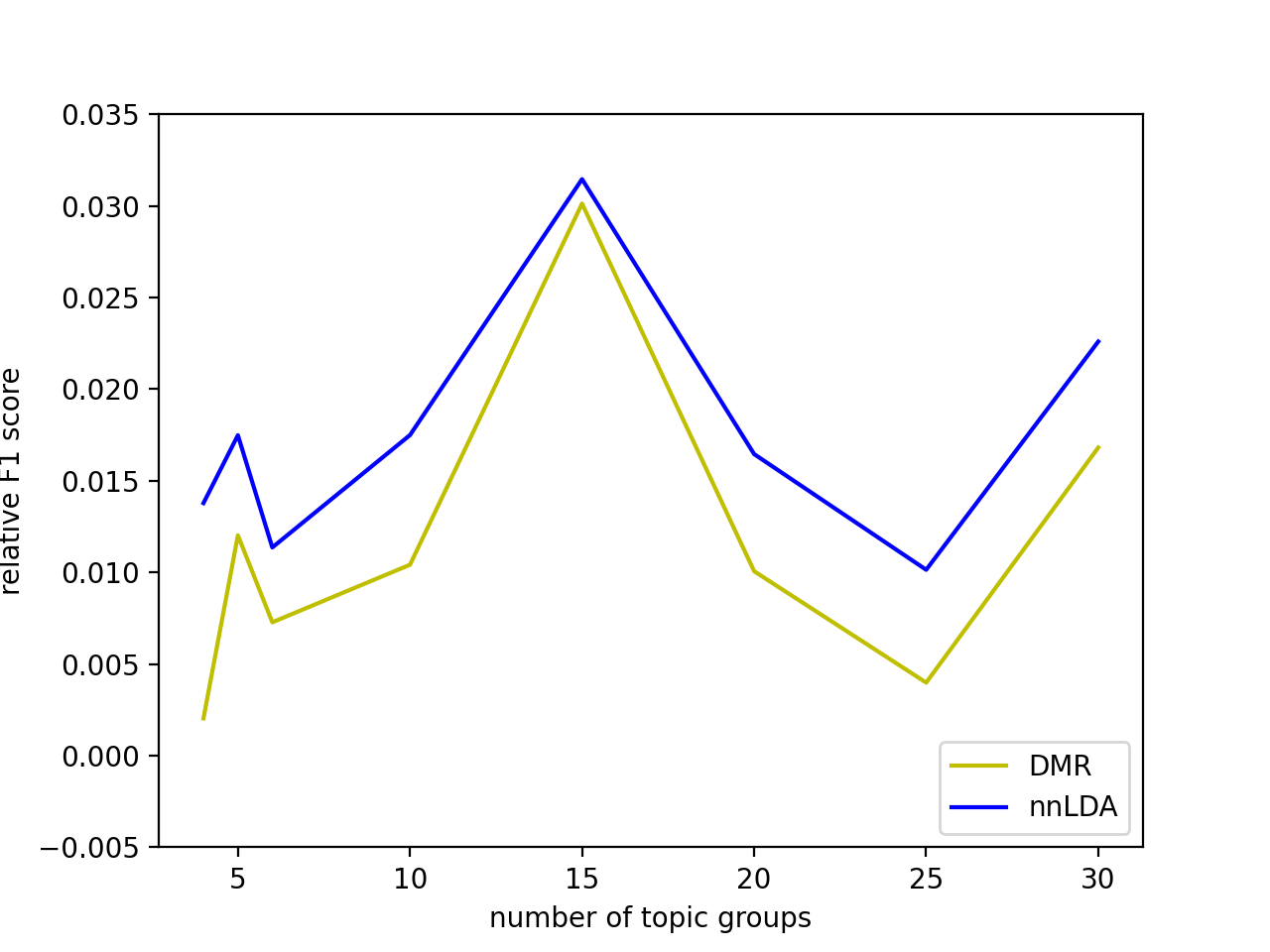}
  \captionof{figure}{PTS}
  \label{fig:proj_f1}
\end{minipage}%
\begin{minipage}{.33\textwidth}
  \centering
  \includegraphics[width=.95\linewidth]{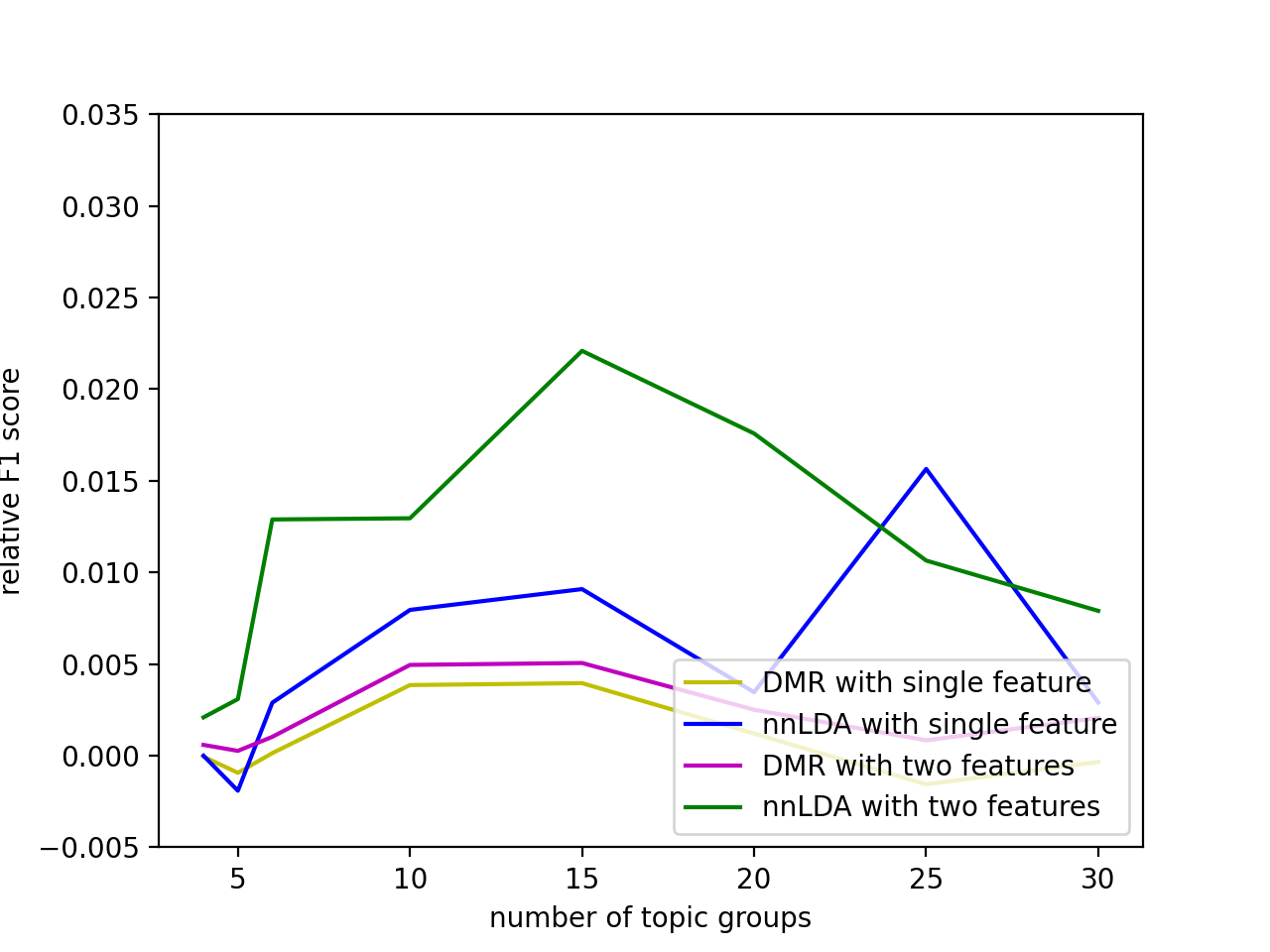}
  \captionof{figure}{WIP}
  \label{fig:whatif_f1}
\end{minipage}%
\begin{minipage}{.33\textwidth}
  \centering
  \includegraphics[width=.95\linewidth]{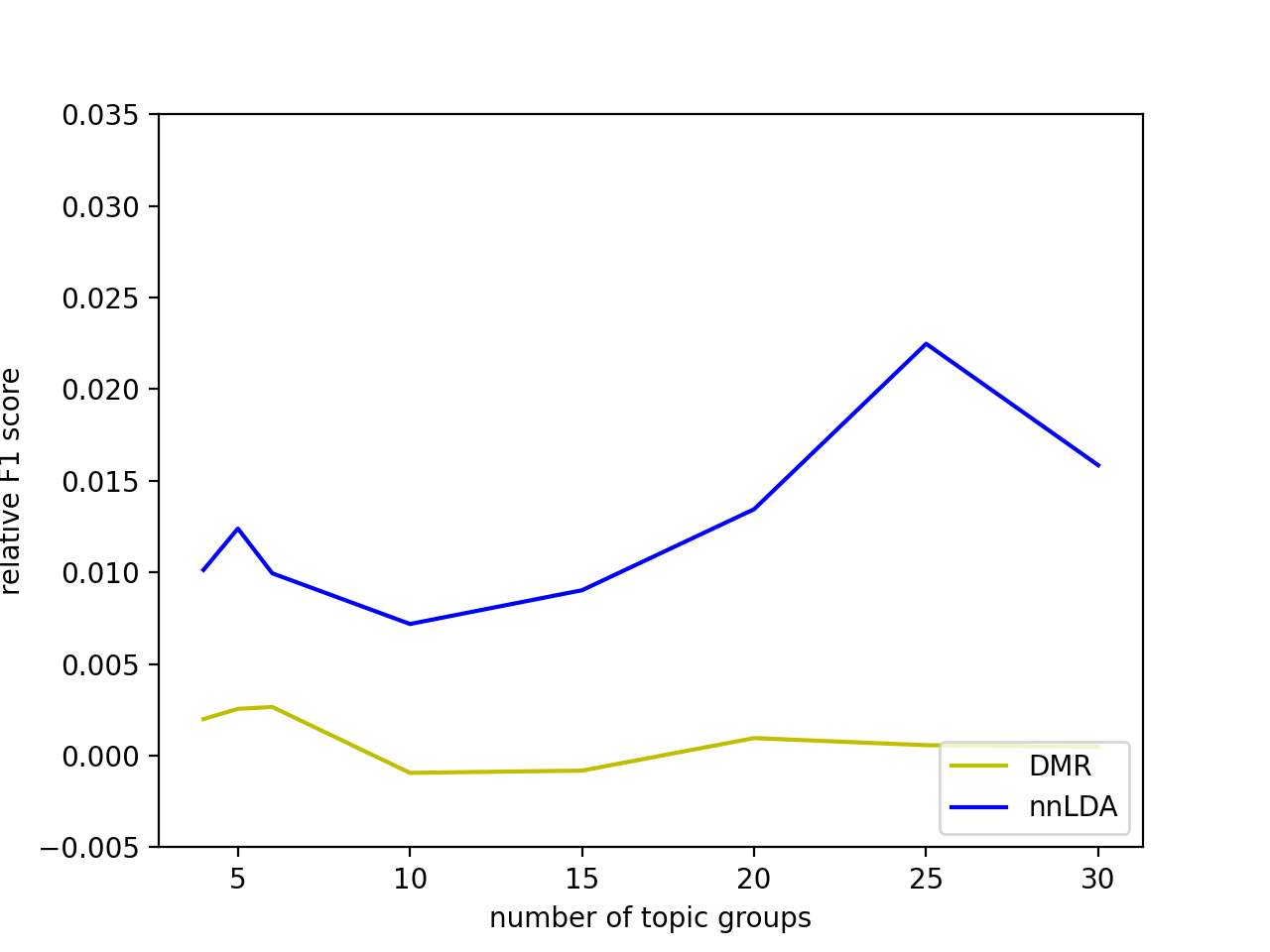}
  \captionof{figure}{DCL}
  \label{fig:TV_f1}
\end{minipage}
\end{figure}

Figures \ref{fig:proj_f1}, \ref{fig:whatif_f1} and \ref{fig:TV_f1} depict the relative F1 scores of DMR and nnLDA with respect to plain LDA on the PTS, WIP, and DCL datasets, respectively. In Figure \ref{fig:proj_f1}, the most distinguishable difference of F1 scores occurs when the number of topic groups is 15, where nnLDA has a gap of 0.032. In the meanwhile, DMR achieves its best performance at the same point with a gap of 0.030. Moreover, this chart shows that nnLDA outperforms plain LDA and DMR no matter what the number of topic groups is. In Figure \ref{fig:whatif_f1}, when using the single feature (sector attribution), the biggest gaps of F1 scores happen when the number of topic groups is 15 for DMR and 25 for nnLDA. The biggest gap between nnLDA and plain LDA is 0.016, while the largest gap between DMR and plain LDA is 0.003. Considering models using two features (sector and channel attributions) as the side data, the highest relative F1 score given by nnLDA with two features is 0.022 with 15 topic groups, compared with 0.004 produced by DMR with 10 topic groups. Although plain LDA provides a slightly higher F1 score than nnLDA when applying 5 topic groups, nnLDA outperforms plain LDA and DMR significantly given any other number of topic groups. In Figure \ref{fig:TV_f1}, the highest relative F1 score given by nnLDA is 0.022 with 25 topic groups, compared with 0.003 given by DMR for 6 topic groups. Moreover, this figure shows that nnLDA outperforms plain LDA dramatically whatever the number of topic groups is. 

Therefore, nnLDA performs better than plain LDA and DMR when predicting the rating given customer’s comments and product information in all datasets.

\subsubsection{Comment Generation}
In this section, we compare the comments generated by nnLDA with plain LDA and DMR. We set the number of topic groups to be 5 since all of plain LDA, DMR and nnLDA have relatively low perplexity scores based on Figures \ref{fig:proj_perp} and \ref{fig:whatif_perp}, and comparable F1 scores based on Figures \ref{fig:proj_f1} and \ref{fig:whatif_f1} on the PTS and WIP datasets. A comment is generated based on the topic-document probability of the sample and the topic-word distribution. More precisely, for DMR and LDA, the prior $\alpha$ is generated based on the side data (sector) first while $\alpha$ is fixed in plain LDA. Next, a comment is created by selecting the top words which have the highest score computed by adding the products of the topic-document probability and topic-word for each word. Then, we randomly pick 50 comments that contain a certain level of information, for example, we rule out comments like \say{N/A.} Meanwhile, in order to evaluate the quality of comment generation, we employed 50 PhD students. Each one of them assessed a pair of comments (one based on plain LDA or DMR, and the other one based on nnLDA) for the same side data and they provided an assessment as to which one is better.

\begin{table}[H]
\centering
\begin{tabular}{|l|r|r|}
\hline  
& \multicolumn{2}{c|}{{ Number of generated comments}} \\ \hline
 & PTS         & WIP         \\ \hline\hline
plain LDA < nnLDA  & \hspace{2.3cm} 15    & 22   \\ \hline
plain LDA > nnLDA & 11   &  9      \\ \hline
plain LDA $\sim$ nnLDA   & 24 & 19  \\ \hline\hline
DMR < nnLDA    & 16  & 20   \\ \hline
DMR > nnLDA   & 11   & 10   \\ \hline
DMR $\sim$ nnLDA  & 23  & 20\\ \hline
\end{tabular}
\caption{Comparison of the generated comments on different datasets}\label{tb10:word generate}
\end{table}

The upper left three values in Table \ref{tb10:word generate} show the comparison of the generated comments given by plain LDA and nnLDA on the PTS dataset. Based on the table, among all these 50 samples, nnLDA generates more accurate comments in 15 samples, while plain LDA does better in 11 samples, and the two are tied for the remaining 24 samples. The lower left three values in Table \ref{tb10:word generate} show the comparison of the generated comments given by DMR and nnLDA on the PTS dataset. Based on the table, among all these 50 samples, nnLDA generates more accurate comments in 16 samples, while DMR does better in 11 samples, and the two are tied for the remaining 23 samples. On the PTS dataset, nnLDA generates in $\frac{15-11}{50}=8\%$ more reasonable comments compared to plain LDA, and in $\frac{16-11}{50}=10\%$ more comparing to DMR.

The right column in Tables \ref{tb10:word generate} shows the comparison of the generated comments given by plain LDA and nnLDA, and DMR and nnLDA on the WIP dataset, respectively. The observations and conclusions are similar. Furthermore, the advantage in number is more obvious on the WIP dataset, i.e. the improvement of nnLDA compared to plain LDA is as large as $\frac{22-9}{50}=26\%$ and the improvement from DMR to nnLDA is $\frac{20-10}{50}=20\%$. Therefore, taking generated comments into consideration, nnLDA generates more reasonable comments than plain LDA and DMR for both small and medium-sized datasets.

\newpage
\bibliographystyle{plain}
\bibliography{references}

\newpage
\section{Appendix}
\subsection*{A \tab Probability Distribution of LDA}
\label{eq:lda}
Given the generative process of LDA, which is formally presented in \cite{blei2003latent}, we obtain the marginal distribution of a document $d=\textbf{w}$ with text only as
\begin{align*}
P_2(\textbf{w}\mid\alpha,\beta)
&=\int \hat{p}(\theta\mid\alpha)\left(\prod_{n=1}^{N}\sum_{z_{k}}\hat{p}(z_{k}\mid \theta)\hat{p}(w_{n}\mid z_{k},\beta)\right)\mathrm{d}\theta\\
&=\int \hat{p}(\theta\mid\alpha)\left(\prod_{n=1}^{N}\sum_{i=1}^{K}\prod_{j=1}^V (\theta_i\beta_{ij})^{w_n^j} \right)\mathrm{d}\theta,
\end{align*}
which in turn yields
\begin{align*}
P_2(D\mid\alpha,\beta)
&=\mathbb{E}\left[\int \hat{p}(\theta_d\mid\alpha)\left(\prod_{n=1}^{N}\sum_{z_{d_k}}\hat{p}(z_{d_k}\mid \theta_d)\hat{p}(w_{d_n}\mid z_{d_k},\beta)\right)\mathrm{d}\theta_d\right]\\
&=\mathbb{E}\left[\int \hat{p}(\theta_d\mid\alpha)\left(\prod_{n=1}^{N}\sum_{i=1}^{K}\prod_{j=1}^V (\theta_i\beta_{ij})^{w_n^j} \right)\mathrm{d}\theta_d\right],
\end{align*}
where $\hat{p}(\theta_d\mid\alpha)=p(\theta_d\mid\alpha)$. 
\subsection*{B \tab Proof of Theorem \hyperref[thm:1]{1}}\label{pf:thm1}
\begin{proof}
By finite sample expressivity of $g(\gamma;\cdot)$, there exists a model with parameters $\gamma_1$ such that
\begin{align*}
g(\gamma_1;\textbf{s})=\alpha^*,
\end{align*}
which in turn yields
\begin{align*}
\tilde{\tilde{p}}(\theta\mid\textbf{s},\gamma_1)=\hat{p}(\theta\mid g(\gamma_1;\textbf{s}))=\hat{p}(\theta\mid\alpha^*).
\end{align*}
Therefore,
\begin{align*}
P_2(D\mid\alpha^*,\beta^*)=\tilde{\tilde{P}}_1(D\mid S,\gamma_1,\beta^*)=P_1(\mu^*, \sigma^*,\gamma_1,\beta^*).
\end{align*}
Since nnLDA also optimizes over the network parameter $\gamma$, we have
\begin{align*}
P_1(D\mid \mu^*, \sigma^*,\gamma^*,\beta^*)\geq P_1(D\mid\mu^*, \sigma^*,\gamma_1,\beta^*),
\end{align*}
and thus,
\begin{align*}
P_1(D\mid\mu^*, \sigma^*,\gamma^*,\beta^*)\geq P_2(D\mid\alpha^*,\beta^*).
\end{align*}
\end{proof}
\subsection*{C \tab Proof of Theorem \hyperref[thm:2]{2}}\label{pf:thm2}
\begin{proof}
Note that
\begin{align}
\label{thm:2-1}
&\frac{P_1(D\mid\mu^*,\sigma^*,\gamma^*,\beta^* )-P_2(D\mid\alpha^*,\beta^*)}{P_2(D\mid\alpha^*,\beta^*)}\nonumber\\
=&
\frac{\mathbb{E}\left[\int \tilde{p}(\theta_d\mid \mu^*,\sigma^*,\gamma^*)\left(\prod_{n=1}^{N}\sum_{z_{d_k}}\tilde{p}(z_{d_k}\mid \theta_d)\tilde{p}(w_{d_n}\mid z_{d_k},\beta^*)\right)\mathrm{d}\theta_d\right]}
{\mathbb{E}\left[\int \hat{p}(\theta_d\mid\alpha^*)\left(\prod_{n=1}^{N}\sum_{z_{d_k}}\hat{p}(z_{d_k}\mid \theta_d)\hat{p}(w_{d_n}\mid z_{d_k},\beta^*)\right)\mathrm{d}\theta_d\right]}\nonumber\\
&\quad\quad\quad-\frac{\mathbb{E}\left[\int \hat{p}(\theta_d\mid\alpha^*)\left(\prod_{n=1}^{N}\sum_{z_{d_k}}\hat{p}(z_{d_k}\mid \theta_d)\hat{p}(w_{d_n}\mid z_{d_k},\beta^*)\right)\mathrm{d}\theta_d\right]}
{\mathbb{E}\left[\int \hat{p}(\theta_d\mid\alpha^*)\left(\prod_{n=1}^{N}\sum_{z_{d_k}}\hat{p}(z_{d_k}\mid \theta_d)\hat{p}(w_{d_n}\mid z_{d_k},\beta^*)\right)\mathrm{d}\theta_d\right]}.
\end{align}
Since
\begin{align*}
&\tilde{p}(\theta_d\mid \mu^*,\sigma^*,\gamma^*)\left(\prod_{n=1}^{N}\sum_{z_{d_k}}\tilde{p}(z_{d_k}\mid \theta_d)\tilde{p}(w_{d_n}\mid z_{d_k},\beta^*)\right)\\
=&\tilde{p}(\theta_d\mid\mu^*,\sigma^*,\gamma^*)\left(\prod_{n=1}^{N}\tilde{p}(w_{d_n}\mid \theta_d,\beta^*)\right)=\prod_{n=1}^{N}\tilde{p}(w_{d_n}\mid \gamma^*,\beta^*, \mu^*,\sigma^*)
\end{align*}
and
\begin{align*}
&\hat{p}(\theta_d\mid\alpha^*)\left(\prod_{n=1}^{N}\sum_{z_{d_k}}\hat{p}(z_{d_k}\mid \theta_d)\hat{p}(w_{d_n}\mid z_{d_k},\beta^*)\right)\\
=&\hat{p}(\theta_d\mid\alpha^*)\left(\prod_{n=1}^{N}\hat{p}(w_{d_n}\mid \theta_d,\beta^*)\right)=\prod_{n=1}^{N}\hat{p}(w_{d_n}\mid \alpha^*,\beta^*),
\end{align*}
equation (\ref{thm:2-1}) could be further simplified as
\begin{align*}
&\frac{P_1(D\mid\mu^*,\sigma^*,\gamma^*,\beta^*)-P_2(D\mid\alpha^*,\beta^*)}{P_2(D\mid\alpha^*,\beta^*)}\\
=& \frac{\mathbb{E}\left[\int\prod_{n=1}^{N}\tilde{p}(w_{d_n}\mid \mu^*,\sigma^*,\gamma^*,\beta^*)\mathrm{d}\theta_d\right]-
\mathbb{E}\left[\int\prod_{n=1}^{N}\hat{p}(w_{d_n}\mid \alpha^*,\beta^*)\mathrm{d}\theta_d\right]}
{\mathbb{E}\left[\int\prod_{n=1}^{N}\hat{p}(w_{d_n}\mid \alpha^*,\beta^*)\mathrm{d}\theta_d\right]}\\
\geq& 
\frac{\mathbb{E}\left[\int C\prod_{n=1}^{N}\hat{p}(w_{d_n}\mid \alpha^*,\beta^*)\mathrm{d}\theta_d\right]-
\mathbb{E}\left[\int\prod_{n=1}^{N}\hat{p}(w_{d_n}\mid \alpha^*,\beta^*)\mathrm{d}\theta_d\right]}
{\mathbb{E}\left[\int\prod_{n=1}^{N}\hat{p}(w_{d_n}\mid \alpha^*,\beta^*)\mathrm{d}\theta_d\right]}\\
=&
\frac{C\cdot \mathbb{E}\left[\int \prod_{n=1}^{N}\hat{p}(w_{d_n}\mid \alpha^*,\beta^*)\mathrm{d}\theta_d\right]-
\mathbb{E}\left[\int\prod_{n=1}^{N}\hat{p}(w_{d_n}\mid \alpha^*,\beta^*)\mathrm{d}\theta_d\right]}
{\mathbb{E}\left[\int\prod_{n=1}^{N}\hat{p}(w_{d_n}\mid \alpha^*,\beta^*)\mathrm{d}\theta_d\right]}=C-1.
\end{align*}

\end{proof}
\end{document}